\documentclass[10pt,twocolumn,letterpaper]{article}

\usepackage[pagenumbers]{cvpr} 

\definecolor{cvprblue}{rgb}{0.21,0.49,0.74}
\usepackage[pagebackref,breaklinks,colorlinks,citecolor=cvprblue]{hyperref}

\def\paperID{5143} 
\def\confName{CVPR}
\def\confYear{2025}

\title{
\includegraphics[height=1.2em]{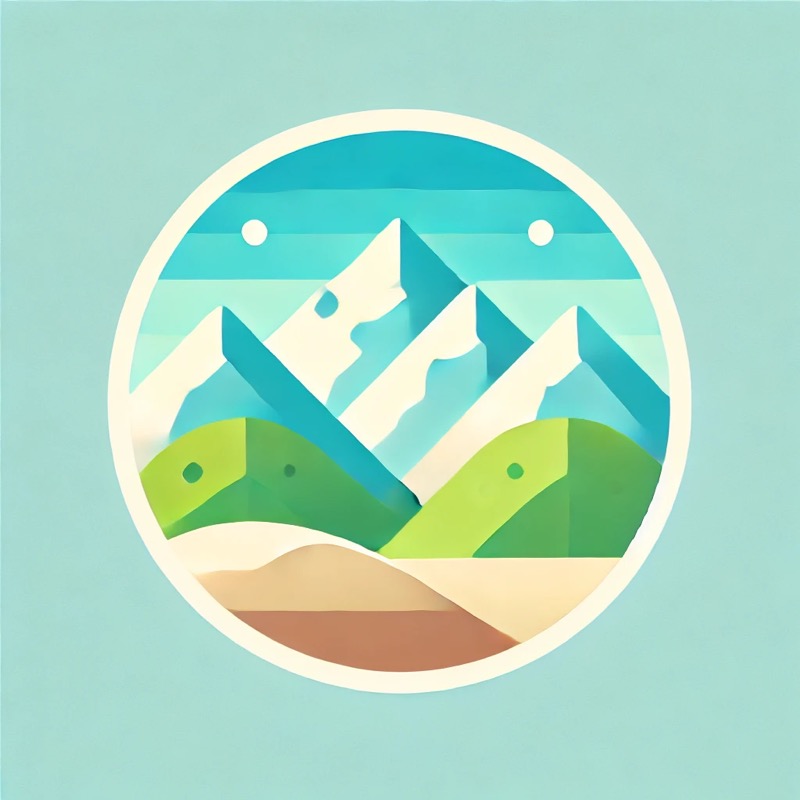} 
RIGI: Rectifying Image-to-3D Generation Inconsistency\\ via Uncertainty-aware Learning}

\author{Jiacheng Wang$^{1}$, \; Zhedong Zheng$^{2}$, \; Wei Xu$^{1}$\footnotemark[2], \; Ping Liu$^{3}$\footnotemark[2] \\
$^1$EIC, Huazhong University of Science and Technology\\
$^2$FST and ICI, University of Macau\\
$^3$CSE, University of Nevada, Reno\\
}

\begin{document}
\twocolumn[{
\renewcommand\twocolumn[1][]{#1}
\maketitle
\begin{center}
    \captionsetup{type=figure}
    \vspace{-.15in}
    \includegraphics[width=0.95\textwidth]{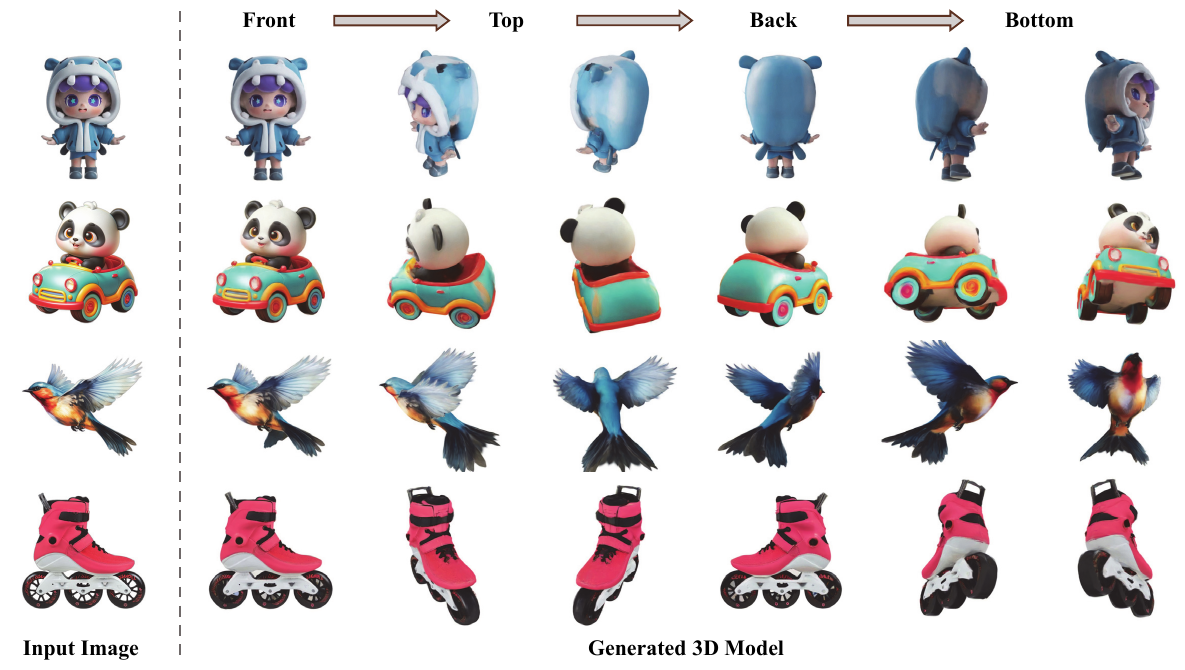}
    \vspace{-.1in}
    \captionof{figure}{Given an input image, our method mitigates the impact of inconsistencies between generated dense frames on 3D asset optimization, reducing edge artifacts and floats while producing visually impressive 3D objects. We sample six rendered images uniformly across an azimuth range of 0 to 360°, with elevations following a sine function with a 30° amplitude, effectively capturing \textbf{front, top, back, and bottom} perspectives, which are crucial to real-world applications, yet often overlooked by most existing methods.}
    \label{fig:vis}
\end{center}
}]
{\renewcommand{\thefootnote}{\fnsymbol{footnote}}
\footnotetext[2]{\;denotes corresponding authors.}}
\begin{abstract}
Given a single image of a target object, image-to-3D generation aims to reconstruct its texture and geometric shape. 
Recent methods often utilize intermediate media, such as multi-view images or videos, to bridge the gap between input image and the 3D target, thereby guiding the generation of both shape and texture. 
However, inconsistencies in the generated multi-view snapshots frequently introduce noise and artifacts along object boundaries, undermining the 3D reconstruction process.
To address this challenge, we leverage 3D Gaussian Splatting (3DGS) for 3D reconstruction, and explicitly integrate uncertainty-aware learning into the reconstruction process. 
By capturing the stochasticity between two Gaussian models, we estimate an uncertainty map, which is subsequently used for uncertainty-aware regularization to rectify the impact of inconsistencies. 
Specifically, we optimize both Gaussian models simultaneously, calculating the uncertainty map by evaluating the discrepancies between rendered images from identical viewpoints.
Based on the uncertainty map, we apply adaptive pixel-wise loss weighting to regularize the models, reducing reconstruction intensity in high-uncertainty regions. 
This approach dynamically detects and mitigates conflicts in multi-view labels, leading to smoother results and effectively reducing artifacts.
Extensive experiments show the effectiveness of our method in improving 3D generation quality by reducing inconsistencies and artifacts. 
More visual results can be found \href{https://rigi3d.github.io/}{here}.
\end{abstract}   
\section{Introduction}
\label{sec:intro}
Image-to-3D generation aims to create 3D objects with corresponding shapes and textures from a single-view image, significantly reducing manual modeling costs and accelerating 3D creation. 
Recent advancements in diffusion models \cite{ldm, imagen, svd} have led to approaches~\cite{dreamfusion, zero123, one2345, magic123, fourier123} that leverage the generative capabilities of pre-trained 2D diffusion models for 3D generation.
However, these 2D diffusion models lack intrinsic 3D perception, often resulting in geometric inconsistencies between generated views.
This limitation compromises the accuracy of the resulting 3D assets, particularly when handling multiple perspectives.

To achieve high consistency in image-to-3D generation, recent research has focused on modifying diffusion models to improve geometric alignment across multi-view images. 
For instance, one line of methods~\cite{mvdream, imagedream, syncdreamer, zero123++, harmonyview, era3d} applies global self-attention to integrate multi-view information, but this increases computational costs, limiting both image resolution and the number of feasible viewpoints.
Alternatively, another line of methods~\cite{im-3d, videomv, v3d, sv3d, hi3d} employs video diffusion models to maintain spatio-temporal consistency, enabling high-resolution, spatially coherent frames that are then optimized into high-quality 3D assets through a reconstruction-based approach.

\begin{figure}[!t]
\centerline{\includegraphics[width=0.48\textwidth]{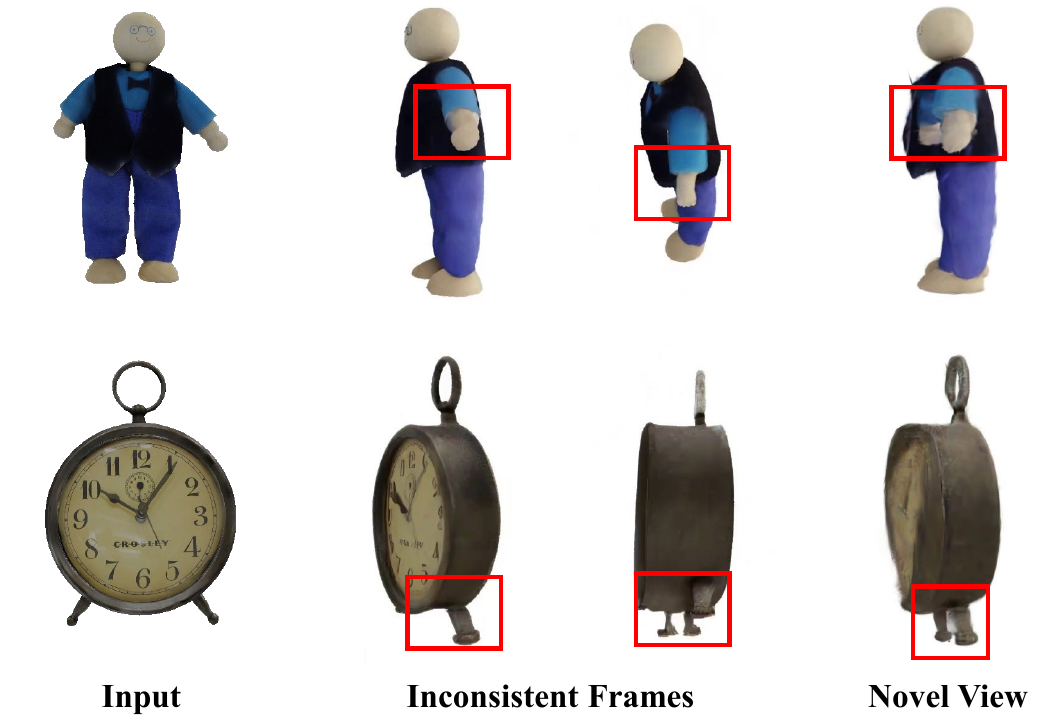}}
\caption{
Prevailing Image-to-3D methods typically adopt the synthesized video as an intermediate representation to guide the 3D object generation.
However, the frame-to-frame inconsistencies can lead to incorrect geometry and artifacts in the 3D assets.
In this example, red bounding boxes highlight extra toy arms and clock legs, which represent common failures in the generation process.
}
\label{fig:intro}
\end{figure}

Despite these advancements, inconsistencies between generated frames still pose significant challenges, introducing artifacts and errors in the 3D generation process, which is illustrated in Figure \ref{fig:intro}.
Pseudo-labels from different viewpoints may exhibit varying geometric structures or textural details within the same 3D region, causing conflicts during subsequent optimization.

To address these issues, we introduce uncertainty-aware learning into the optimization process of 3D assets.
Given its efficient training and high-quality rendering, we use 3DGS~\cite{3dgs} as our 3D representation.
Our approach consists of two key steps: uncertainty estimation and uncertainty regularization.
In the first step, we simultaneously optimize two Gaussian models and model uncertainty by capturing the stochastic differences between them. 
The uncertainty map is estimated by calculating the absolute difference between the rendered images of the two models.
In the second step, we conduct uncertainty regularization based on the estimated uncertainty map.
This step dynamically adjusts pixel-wise loss weights, reducing the impact of inconsistent pseudo-labels in high-uncertain regions.
Our approach dynamically detects pseudo-label inconsistencies during optimization and progressively alleviates their negative impact on 3D assets generation. 
Extensive experiments show that it mitigates artifacts and improves 3D generation quality both quantitatively and qualitatively.
In summary, our contributions are as follows: 
\begin{itemize}
    \item We introduce uncertainty-aware learning into 3D Gaussian Splatting, modeling the uncertainty between generated pseudo-labels by leveraging the variations between two concurrently optimized Gaussian models. 
    We apply uncertainty regularization based on  estimated uncertainty maps to mitigate conflicts in generated pseudo-frames.
    \item Extensive experiments show that our approach, combined with a multi-view video diffusion model, produces 3D assets of high quality, as evidenced by both quantitative and qualitative analyses.
\end{itemize}

\section{Related Work}
\label{sec:related_work}

\noindent\textbf{Image-to-3D Generation.}
Image-to-3D generation aims to create accurate 3D assets from a single 2D image, a challenging task requiring reliable modeling of unseen views.  
Early methods focused primarily on single-view 3D reconstruction~\cite{occupancy_net, pixel2mesh, grf, topologically,zhang2023multi}, while recent advancements have shifted towards employing image-based 3D generative models with diverse 3D representations~\cite{shap-e, point-e, 3dgen, direct3d, rodin}, enabling the generation of more complex 3D assets.

While traditional 3D generation methods require high-quality 3D data, limiting their generalization, large-scale 2D diffusion models~\cite{ldm, imagen, svd} have shown success in generating high-quality images and videos, inspiring 2D-to-3D lifting approaches.
DreamFusion~\cite{dreamfusion} and Zero123~\cite{zero123} leverage pre-trained 2D diffusion models to optimize 3D assets and synthesize new views, demonstrating strong 3D consistency. 
However, these methods remain computationally expensive. 
To address this, recent research has focused on more efficient feed-forward approaches, particularly large reconstruction models (LRMs)~\cite{lrm, triposr, sf3d}, which map image features into 3D triplane space for faster 3D asset creation. 
Subsequent works~\cite{mvdream, imagedream, syncdreamer, zero123++, harmonyview, era3d,yi2023progressive} have integrated multi-view diffusion models, improving both geometry and texture quality.
The methods leveraging temporal consistency in video models~\cite{im-3d, videomv, v3d, sv3d, hi3d} enhance 3D reconstruction by first generating dense video frames and then performing 3D reconstruction.

In this paper, we follow a two-stage approach that yields visually impressive results with detailed textures, leveraging the ability of video models to generate dense and high-quality frames. 
However, the multi-view frames generated by these methods often exhibit inconsistencies in both geometry and texture.
Addressing the impact of these inconsistencies on subsequent 3D asset optimization remains an important challenge and the focus of our work.

\noindent\textbf{Uncertainty-aware Learning.} With the advancement of deep learning, there is a growing focus on improving the reliability and interpretability of the model. 
Estimating the model uncertainty not only improves the interpretability but also provides a quantitative measure of output confidence.
Early work~\cite{uncertainties} categorizes uncertainty into two types, \ie, epistemic uncertainty and aleatoric uncertainty. Epistemic uncertainty, also known as model uncertainty, refers to the variability in model weights trained on the same dataset. 
Pioneering methods such as Bayesian networks~\cite{bayesian, bayesian2}, dropout~\cite{dropout, dropout2}, and adding Gaussian noise~\cite{yu2019robust,chen2024composed} exploit inherent randomness in neural networks to estimate this uncertainty, typically by modeling the variance in weight distributions.
Other approaches~\cite{uncert_net_1, uncert_net_2, uncert_net_3, zhang2024harnessing} explicitly model uncertainty through an auxiliary branch, though at the cost of higher training expenses and potential accuracy trade-offs.
Aleatoric uncertainty, on the other hand, represents noise in observations, including both input data and annotations. 
Many approaches~\cite{uncertainties, aleatoric_1, aleatoric_2} model aleatoric uncertainty to identify noise in inputs or annotations, using a dynamic uncertainty-aware loss to stabilize training and improve the final results.

Uncertainty estimation has been widely explored in the 3D domain, particularly within Neural Radiance Fields (NeRF) and 3D Gaussian Splatting (3DGS). 
Uncertainty in NeRF models arises from factors such as variations in camera models, lighting conditions~\cite{nerf_1, nerf_2}, occlusions, and sparse viewpoints~\cite{sparsenerf_1, sparsenerf_2, sparsenerf_3}. 
Approaches such as Bayesian reparameterization~\cite{activenerf, sparsenerf_2, sparsenerf_3} and volume rendering~\cite{active, activermap} model uncertainty by quantifying the variability in weight distributions or density along rays. 
Fisher information has also been applied~\cite{fisherrf, bayes_ray} to quantify uncertainty in rendered views.

Recently, 3DGS has garnered significant attention for its high-quality reconstruction and efficient rendering, incorporating uncertainty estimation to improve output confidence. 
Studies~\cite{modeling_uncert, source_uncert} have investigated modeling uncertainty within 3DGS, facilitating effective quantification of output confidence and alleviating the impacts of noise, occlusion, and imprecise camera poses on the reconstruction process.
Additionally, uncertainty information has been used for next-view selection~\cite{next_view_1, next_view_2}, optimizing the reconstruction process by identifying beneficial viewpoints, thus reducing the need for extensive scene capture.
More recent works~\cite{uncert_3dgs_1, uncert_3dgs_2} have integrated uncertainty-aware learning into training, dynamically adjusting pixel contributions to minimize noise in uncertain regions.

In this work, we tackle the challenging problem of Image-to-3D generation, focusing on mitigating noise and inconsistencies across synthesized multi-view frames.
Our approach introduces a dynamic, uncertainty-aware mechanism that adapts pixel-wise loss weights based on the estimated uncertainty,  enhancing the robustness of the generation process. 
This method significantly reduces visual artifacts and distortion in generated 3D assets.
\section{Method}

\begin{figure*}[!t]
\centerline{\includegraphics[width=0.98\textwidth]{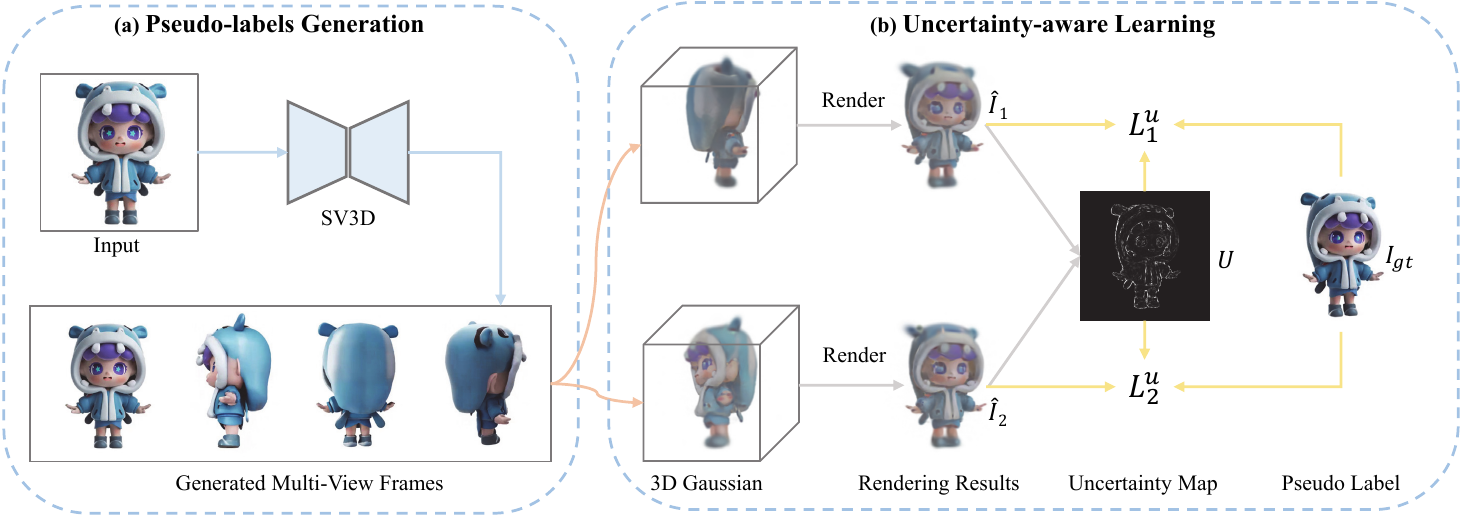}}
\caption{\textbf{Overview of our pipeline.} Firstly, we use SV3D~\cite{sv3d} to generate multiple videos with a wide range of viewpoints, which serve as pseudo-labels for 3D asset optimization. Next, we introduce uncertainty-aware learning, estimating an uncertainty map by leveraging the stochasticity of two simultaneously optimized Gaussian models.
Finally, we apply uncertainty-aware regularization to mitigate the impact of inconsistencies in the generated pseudo-labels, resulting in high-quality and visually impressive 3D assets.}
\label{fig:pipe}
\end{figure*}

Our pipeline, as shown in Figure \ref{fig:pipe}, takes a reference image as input and outputs 3D assets.
We adopt a two-stage approach: first, a multi-view video diffusion model generates dense, high-quality frames, which serve as pseudo-labels to guide 3D asset optimization; second, uncertainty-aware learning is applied to optimize the 3D assets, improving robustness and reducing artifacts in the final output.
In Section \ref{sec: preliminary}, we introduce 3D Gaussian Splatting  and the video diffusion models that form the foundation of our approach.
In Section \ref{sec: uncert estimation} and \ref{sec: uncert regularization}, we describe the integration of uncertainty-aware learning into the 3D asset optimization process, encompassing both uncertainty estimation and regularization. 
Finally, in Section \ref{sec: optimization strategy}, we discuss additional optimization strategies used to further enhance visual quality throughout the optimization process.

\subsection{Preliminary}
\label{sec: preliminary}

\textbf{3D Gaussian Splatting.} 3DGS~\cite{3dgs} is a point-based explicit 3D representation composed of a set of learnable Gaussian points. 
Each Gaussian point is parameterized by the center position $\mu_i \in \mathbb{R}^{3}$, scaling $s_i \in \mathbb{R}^{3}$, rotation $r_i \in \mathbb{R}^{4}$, color $c_i \in \mathbb{R}^{3}$, spherical harmonic (SH) coefficients $h_i \in \mathbb{R}^{3\times (k+1)^2}$ up to order k, and opacity $\sigma_i \in \mathbb{R}$. The Gaussian model can be queried as:
\begin{equation}
    G(x)=e^{-\frac{1}{2}(x-\mu)^{\top} \boldsymbol{\Sigma}^{-1}(x-\mu)},\ \Sigma_i=R_iS_iS_i^TR_i^T,
\end{equation}
where $x$ is a given 3D position, $\boldsymbol{\Sigma}$ is the covariance matrix, $S_i$ and $R_i$ are the scaling and rotation matrices derived from $s_i$ and $r_i$, respectively. 

To render a 2D image from a specific camera pose, the color of each pixel is determined by $\alpha$-blending of the sorted Gaussian points:
\begin{equation}
    C=\sum_{i=1}^Nc_i\alpha_i\prod_{j=1}^{i-1}\left(1-\alpha_j\right),
\end{equation}
where $c_i$ is the color and $\alpha_i$ represent the projected opacity.

The Gaussian model is trained using a reconstruction loss function: $\mathcal{L}=(1-\lambda)\mathcal{L}_{1}+\lambda\mathcal{L}_{\mathrm{D\text{-}SSIM}}$, where $\lambda$ is  empirically set to $0.2$. 
Starting from a sparse point cloud, the model applies a densification process that duplicates and splits high-gradient Gaussian points, complemented by a pruning strategy to eliminate non-essential points.
In this work, we adopt 3DGS as our 3D representation due to its efficient training and high-quality rendering capabilities.

\noindent\textbf{Video Diffusion Model.} 
Video diffusion models~\cite{svd, align_latents} are typically built upon pre-trained 2D image diffusion models~\cite{ddpm, ddim, ldm}, enabling the generation of spatially and temporally consistent video sequences by denoising multiple frames simultaneously.
A representative approach is Stable Video Diffusion (SVD)~\cite{svd}, which consists of an encoder $\varepsilon$, a denoising U-Net $\epsilon_\theta$, and a decoder $\mathcal{D}$.
SVD achieves high-quality and consistent video generation in recent works. 
Given a condition image $c$ and an initial sequence of random noise $x_T$, the denoising U-Net estimates the added noise at timestep $t$.
The noise scheduler~\cite{edm} progressively removes noise at each timestep to produce $x_{t\text{-}1}$,
which can be written as:
\begin{equation}
    x_{t\text{-}1}=\Phi\left(\epsilon_\theta\left(x_{t}; t, c\right), t, x_t\right),
\end{equation}
where $\Phi$ denotes the noise scheduler and $c$ represents the condition embedding.
After $T$ denoising steps, a high-quality sequence of $N$ video frames is generated.

Building on the spatiotemporal consistency of SVD, SV3D~\cite{sv3d} adapts the denoising U-Net to condition on camera pose, offering precise control over viewpoints and improving multi-view consistency in image-to-3D generation:
\begin{equation}
    x_{t\text{-}1}=\Phi\left(\epsilon_\theta\left(x_{t}; t, c, a, e\right), t, x_t\right),
\end{equation}
where $a$ and $e$ denote the azimuth and elevation angles, respectively.
With an input image, SV3D can generate $N$ frames from different viewpoints, supporting both static and dynamic camera orbits. 
In this study, we employ SV3D to generate multi-view frames as pseudo-labels for 3D assets optimization, leveraging its state-of-the-art performance in geometric and texture consistency, along with its ability to control dynamic camera poses.

\subsection{Uncertainty Estimation}
\label{sec: uncert estimation}
In Image-to-3D generation, uncertainty arises from two main sources: epistemic uncertainty, due to the limited information from a single image, and aleatoric uncertainty, due to inherent noise or inconsistencies in the input data. 
Epistemic uncertainty leads to variations in the 3D assets, particularly in unobserved regions, which is mitigated by using a multi-view video diffusion model to generate dense frames from multiple viewpoints. 
Aleatoric uncertainty results from visual overlaps between pseudo frames, introducing inconsistencies in geometry and texture that can cause artifacts during optimization process.

As discussed, uncertainty estimation is crucial for mitigating the impact of noisy pseudo-labels on 3D asset optimization. 
To estimate uncertainty, we model the discrepancies between two simultaneously optimized Gaussian models. 
This approach leverages the inherent randomness in the training process: when optimized with the same data, 3D assets will naturally exhibit variations.
These variations indicate the degree of uncertainty in the generated pseudo-labels. 
Specifically, regions with higher uncertainty will show greater variability, resulting in more pronounced differences between the models, while regions with lower uncertainty will exhibit less variation. 
By capturing these differences, we accurately estimate the inconsistency in the noisy labels, and in turn, guide the optimization process to improve the stability and accuracy of the 3D generation.

Specifically, given a set of generated multi-view labels $\mathcal{I}^{gt}=\left\{I_{i}^{gt}\right\}_{i=1}^{M}$ corresponding to various camera poses $\mathcal{P}=\left\{P_{i}\right\}_{i=1}^{M}$, we simultaneously optimize two Gaussian models, $\mathcal{G}_{1}$ and $\mathcal{G}_{2}$, where $M$ denotes the number of frames. 
During each optimization step, we randomly sample a camera pose $P_{i} \in \mathcal{P}$, and render both Gaussian models from the corresponding viewpoint to obtain the rendered images $\hat{I_{1}}$ and $\hat{I_{2}}$. 
Then the uncertainty of the Gaussian models under a given camera pose is approximated by the difference between the rendered images $\hat{I_{1}}$ and $\hat{I_{2}}$:
\begin{equation}
    U = |\hat{I}_{1}-\hat{I}_{2}|.
\end{equation}
Notably, each Gaussian model is randomly initialized, ensuring observable differences between $\mathcal{G}_{1}$ and $\mathcal{G}_{2}$ throughout the optimization.
This variability enables the models to effectively capture and model the uncertainty in the pseudo-labels, guiding the optimization effectively.

\textbf{Why not adopt a learnable approach to model uncertainty?}
For 3DGS, one approach to model uncertainty is to assign a learnable variance property to each Gaussian point, allowing an uncertainty map to be rendered through $\alpha$-blending.
During optimization, the variance parameter is dynamically updated. 
However, directly regressing the uncertainty in this manner may lead to training instability, as the variance of certain Gaussian points may become excessively large or small, hindering the achievement of an optimal result.
In contrast, our approach avoids directly optimizing the variance.
Instead, we model uncertainty by capturing the differences between two Gaussian models, which enhances optimization stability. 
Our experiments show that two Gaussian models are sufficient, and their absolute difference effectively captures uncertainty. 
Based on the estimated uncertainty map, uncertainty-aware regularization effectively mitigates artifacts and floats in 3D assets caused by inconsistencies in pseudo-labels.

\subsection{Uncertainty Regularization}
\label{sec: uncert regularization}
In Image-to-3D generation, inconsistencies in pseudo multi-view frames often cause issues during optimization, especially in regions with high uncertainty. 
The inconsistencies arise when pseudo-labels from different viewpoints conflict, leading to conflicting optimization directions and artifacts or floats in the generated 3D assets. 
To address this, we adjust the pixel-wise loss by incorporating our estimated uncertainty map.
Specifically, we modify the original loss function, which includes a pixel-wise loss term and a D-SSIM term, to account for these inconsistencies:
\begin{equation}
\begin{aligned}
    \mathcal{L}_1^u=\frac{|I_{gt} - \hat{I}_1|}{\exp{(\lambda\cdot U)}}+\lambda\cdot U, \\
    \mathcal{L}_2^u=\frac{|I_{gt} - \hat{I}_2|}{\exp{(\lambda\cdot U)}}+\lambda\cdot U,
\end{aligned}
\end{equation}
where $U$ represents the uncertainty between the two Gaussian models, $\lambda$ controls the strength of the uncertainty regularization.  
By dynamically adjusting the loss based on the uncertainty, we improve the stability and quality of the 3D asset generation, particularly in uncertain regions.

Notably, we amplify the estimated uncertainty map to increase the variation in regularization loss, using an experimentally determined factor $\lambda=5$.
The first term dynamically adjusts the optimation intensity for each pixel in the pseudo-labels, based on the estimated uncertainty map.
Regions with higher inconsistencies receive lower weights,  reducing their influence on the optimization process.
The second term regularizes the uncertainty map, preventing excessive disparity between the two Gaussian models that could result in high uncertainty across all viewpoints.
When the uncertainty map is constant at zero, the regularization term reduces to the standard L1 loss, where each region is optimized with a constant weight. Given the  effectiveness of LPIPS loss~\cite{lpips} in enhancing visual quality, we incorporate it into the reconstruction loss.
The LPIPS loss and the D-SSIM Loss~\cite{ssim} are then formulated as:
\begin{equation}
\begin{aligned}
    \mathcal{L}_{lpips}=\mathrm{LPIPS}(I_{gt}, \hat{I}_1)+\mathrm{LPIPS}(I_{gt}, \hat{I}_2), \\
    \mathcal{L}_{d\text{-}ssim}=\mathrm{D\text{-}SSIM}(I_{gt}, \hat{I}_1)+\mathrm{D\text{-}SSIM}(I_{gt}, \hat{I}_2).
\end{aligned}
\end{equation}
Finally, the total loss is given by:
\begin{equation}
    \mathcal{L}_{total}=(1-\lambda_{s})(\mathcal{L}_1^u+\mathcal{L}_2^u)+\lambda_{s}\mathcal{L}_{d\text{-}ssim}+\lambda_{l}\mathcal{L}_{lpips},
\end{equation}
where $\lambda_{s}$ and $\lambda_{l}$ are empirically set to 0.2 and 0.5.

\textbf{How does uncertainty regularization impact the optimization of 3D assets?} Generated multi-view frames often exhibit inconsistencies in overlapping regions, where pseudo-labels from different viewpoints may conflict in their optimization directions.
In such cases, Gaussian models may densify redundant points to satisfy the pseudo-labels from certain viewpoints, potentially leaving uncovered viewpoints vulnerable to artifacts or floats. By applying uncertainty regularization, regions with high inconsistency are assigned higher uncertainty values.
This reduces the conflicting influence of pseudo-labels, diminishing the need to densify redundant Gaussian points and leading to a smoother outcome in inconsistent regions.

\subsection{Optimization Strategy}
\label{sec: optimization strategy}
To enhance the quality of 3D generation, we apply several optimization techniques along with uncertainty estimation and regularization. Specifically, we integrate Perturbed-Attention Guidance (PAG)~\cite{pag} into the multi-view video diffusion model, improving texture and geometry, particularly in rear-view perspectives. 
Camera poses are sampled across a range of elevation and azimuth angles to ensure robust 3D asset generation from various viewpoints. For optimization, we use multi-scale rendering and progressive sampling strategies to balance training efficiency and visual quality. 
Initially, the optimization focuses on geometric structure, with progressive increases in resolution to capture finer texture details. 
We also progressively introduce frames with varying elevations to stabilize the training process and reduce inconsistencies in geometry initialization.
Additionally, we mitigate redundant white Gaussian points in inconsistent regions by applying a random background color technique, effectively reducing white artifacts. 
These strategies enhance both reconstruction quality and visual consistency, resulting in more impressive 3D outputs.
\section{Experiments}

\begin{table}[!t]
\caption{\textbf{Quantitative Comparison.} Our method achieves superior or comparable results, demonstrating its effectiveness in generating high-quality 3D assets.}
\label{table:quantitative_comparison}
\small
\centering
\begin{tabular}{c|ccc}
\toprule
Methods          & PSNR$\uparrow$    & SSIM$\uparrow$   & LPIPS$\downarrow$  \\
\midrule
DreamGaussian~\cite{dreamgaussian}    & 17.162  & 0.8252 & 0.2039 \\
TriplaneGaussian~\cite{triplanegaussian} & 14.0062 & 0.8161 & 0.2531 \\
LGM~\cite{lgm}              & 14.5874 & 0.8083 & 0.2488 \\
V3D~\cite{v3d}              & 17.1847 & 0.8085 & 0.2055 \\
Hi3D~\cite{hi3d}             & \textbf{17.2559} & 0.8217 & 0.2014 \\
Ours             & 16.9646 & \textbf{0.8346} & \textbf{0.2003} \\
\bottomrule
\end{tabular}
\end{table}

\begin{figure*}[!t]
\centerline{\includegraphics[width=1.0\textwidth]{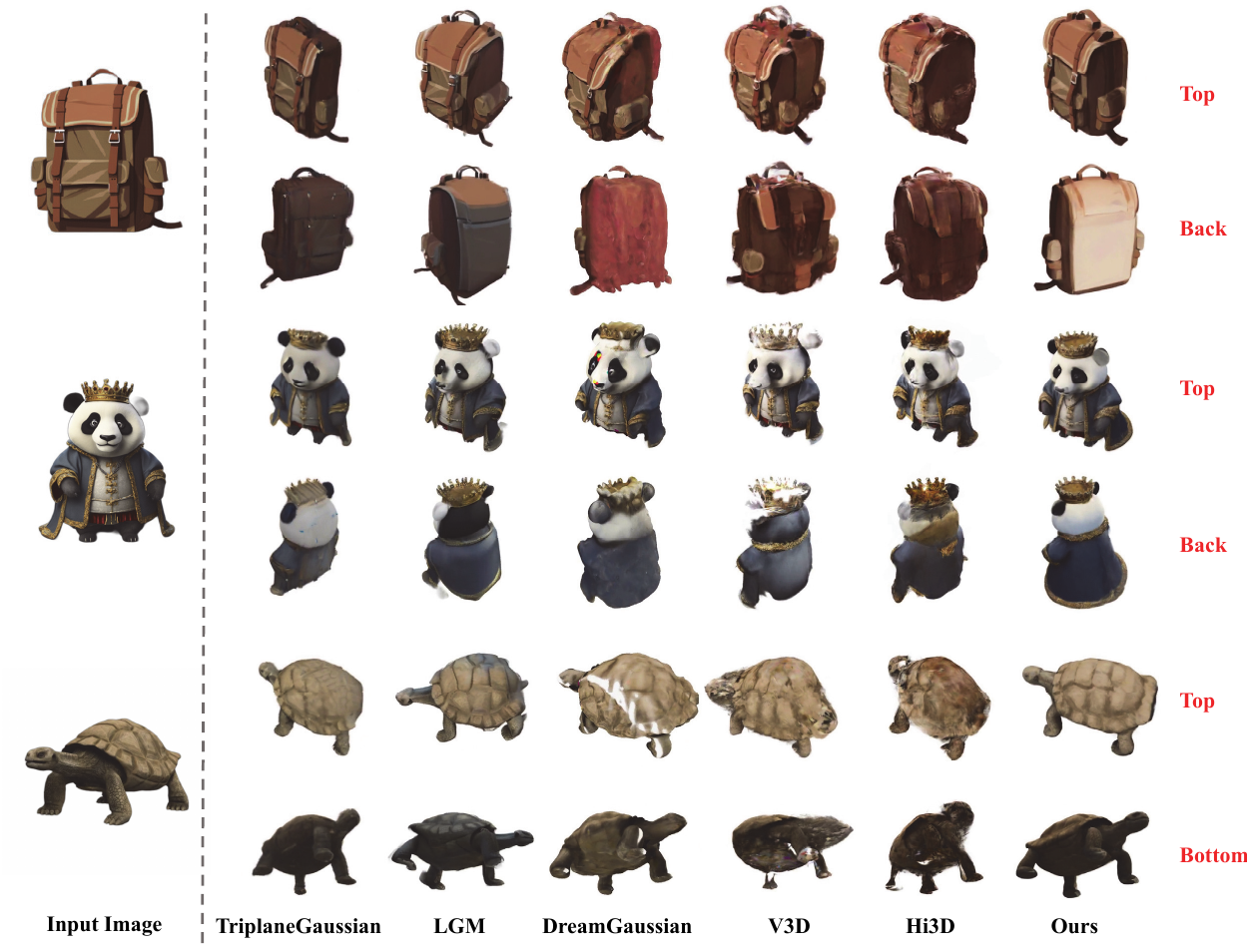}}
\caption{\textbf{Visual Comparison}. 
Here we compare competitive image-to-3D methods, including TriplaneGaussian~\cite{triplanegaussian}, LGM~\cite{lgm}, Dreamgaussian~\cite{dreamgaussian}, V3D~\cite{v3d} and Hi3D~\cite{hi3d}.
We achieve visually impressive results, with high-quality geometric and texture details even from top and bottom perspectives. 
}
\label{fig:visual_cmp}
\end{figure*}

\begin{table*}[!t]
\caption{\textbf{User Study}. 
We curate a set of 30 samples and conduct a user study with 35 participants, each tasked with selecting the top two results that best matched the input image and exhibited the highest visual quality. Our method achieved the highest preference score, demonstrating its capability to produce visually compelling 3D assets.}
\label{table:user_study}
\small
\centering
\setlength{\tabcolsep}{3.5mm}{
\begin{tabular}{c|cccccc}
\toprule
Methods & TriplaneGaussian~\cite{triplanegaussian} & LGM~\cite{lgm} & DreamGaussian~\cite{dreamgaussian} & V3D~\cite{v3d} & Hi3D~\cite{hi3d} & Ours \\
\midrule
Preference$\uparrow$ & 19.81\% & 46.95\% & 25.14\% & 19.24\% & 21.91\% & \textbf{66.95\%} \\
\bottomrule
\end{tabular}}
\end{table*}

\noindent\textbf{Implementation Details.}  For multi-view frames generation, we employ $sv3d\_p$~\cite{sv3d}, which generates frames from various viewpoints along a dynamic orbit. 
We then sample multiple videos with azimuth angles uniformly distributed across 360°, and elevation angles defined by sinusoidal amplitudes of 0°, -20°, and 40°, resulting in 63 frames in total.
For 3D asset optimization, we follow the original setup of 3DGS~\cite{3dgs}, with minor modifications. 
Specifically, the spherical harmonics (SH) degree is set to $0$, and the total optimization iterations is reduced to $5000$. 
During the optimization process, we progressively increase the render radio, beginning at 0.25 and scaling up to 0.5 at 20\% of the total iterations, and reaching 1.0 at 50\% of the total iterations. 
Additionally, we progressively incorporate frames with different elevations throughout the optimization process, at 50\% and 80\% of the total iterations, respectively.
For the camera setup, the field of view (FOV) is configured at 33.8° with a radius of 4.0.

\noindent\textbf{Baselines.}
We selected five image-to-3D generation methods based on 3D Gaussian Splatting~\cite{3dgs} for comparison:
(1) \textbf{DreamGaussian}~\cite{dreamgaussian} is an optimization-based approach that refines 3D assets under the supervision of Zero123~\cite{zero123};
(2) \textbf{TriplaneGaussian}~\cite{triplanegaussian} is an inference-only method that proposes a hybrid triplane-gaussian representation to achieve fast and high-quality 3D reconstruction;
(3) \textbf{LGM}~\cite{lgm} is another inference-only method that reconstructs Gaussian models from generated multi-view images;
(4) \textbf{V3D}~\cite{v3d} is a multi-view video diffusion model that generates dense frames, which are then used as pseudo-labels for 3D asset reconstruction;
(5) \textbf{Hi3D}~\cite{hi3d} employs a two-stage generation paradigm to produce high-resolution multi-view frames, enhancing the generated texture details.

\noindent\textbf{Evaluation Metrics.}
To evaluate the visual quality of the generated assets, we selected 25 objects from the GSO dataset~\cite{gso}, manually choosing front-facing input images. 
We rendered 36 ground truth images with uniformly sampled azimuth angles and randomly sampled elevation angles, ensuring coverage of both top and bottom perspectives of the 3D assets.
We then used PSNR, SSIM, and LPIPS as evaluation metrics to measure the difference between the generated views and the ground truth.

\subsection{Experimental Results}

\noindent\textbf{Qualitative Results.} 
As shown in Figure \ref{fig:visual_cmp}, we provide qualitative comparisons across several approaches, including optimization-based, inference-only, and two-stage methods. 
TriplaneGaussian~\cite{triplanegaussian}, though efficient in generating 3D assets, yields lower-resolution outputs with limited texture detail.
LGM~\cite{lgm} employs an asymmetric U-Net to produce high-resolution 3D objects; however, inconsistencies in the input multi-view images may result in artifacts and floats.
DreamGaussian~\cite{dreamgaussian} leverages SDS Loss for 3D object optimization, but it often generates coarse textures on the back, leading to a disconnect between front and back views.
V3D~\cite{v3d} and Hi3D~\cite{hi3d} employ multi-view video diffusion models to generate dense, high-quality frames, producing 3D objects with detailed textures. 
However, since optimization is performed from a limited set of fixed viewpoints, these methods may overfit to the generated frames, resulting in underdeveloped geometry and texture details from top and bottom perspectives.
Our approach samples videos from diverse viewpoints and uses uncertainty-aware learning to mitigate inconsistencies, resulting in visually impressive 3D outputs.

\noindent\textbf{Quantitative Results.} 
We selected 25 objects from the GSO~\cite{gso} dataset and used SSIM, PSNR, and LPIPS to evaluate the visual quality of the generated 3D objects. 
As shown in Table \ref{table:quantitative_comparison}, we achieve superior or comparable results, demonstrating the effectiveness of our approach in generating high-quality and visually impressive 3D assets.
Specifically, our method performs well on SSIM, indicating excellent structural consistency and effectively mitigating noise, artifacts, and floats in inconsistent areas. 
Additionally, we observe a slight improvement in LPIPS, suggesting good perceptual quality in the generated 3D assets. 
However, we did not achieve the highest PSNR score, slightly trailing behind Hi3D~\cite{hi3d}, which may seem inconsistent with the qualitative results. 
We hypothesize that uncertainty regularization dynamically adjusts pixel-level supervision in the pseudo-labels, prioritizing structural smoothness in inconsistent regions rather than redundant Gaussian points for pixel-level alignment. 
Consequently, this approach may yield slightly lower scores on pixel-wise metrics, which tend to favor precise structural coherence.

\noindent\textbf{User Study.} 
To evaluate visual quality, we curated a set of $30$ samples and conducted a user study with $35$ participants, as summarized in Table \ref{table:user_study}. 
Each participant was asked to select the top two results that best matched the input image and exhibited the highest visual quality, with the total preference score in the table summing to 200\%.
The collected preferences were then analyzed to compare the performance of our method with other state-of-the-art approaches.
As shown in the results, our method was selected more frequently, demonstrating its ability to consistently produce the most visually compelling 3D assets.

\begin{figure}[!t]
\centerline{\includegraphics[width=0.48\textwidth]{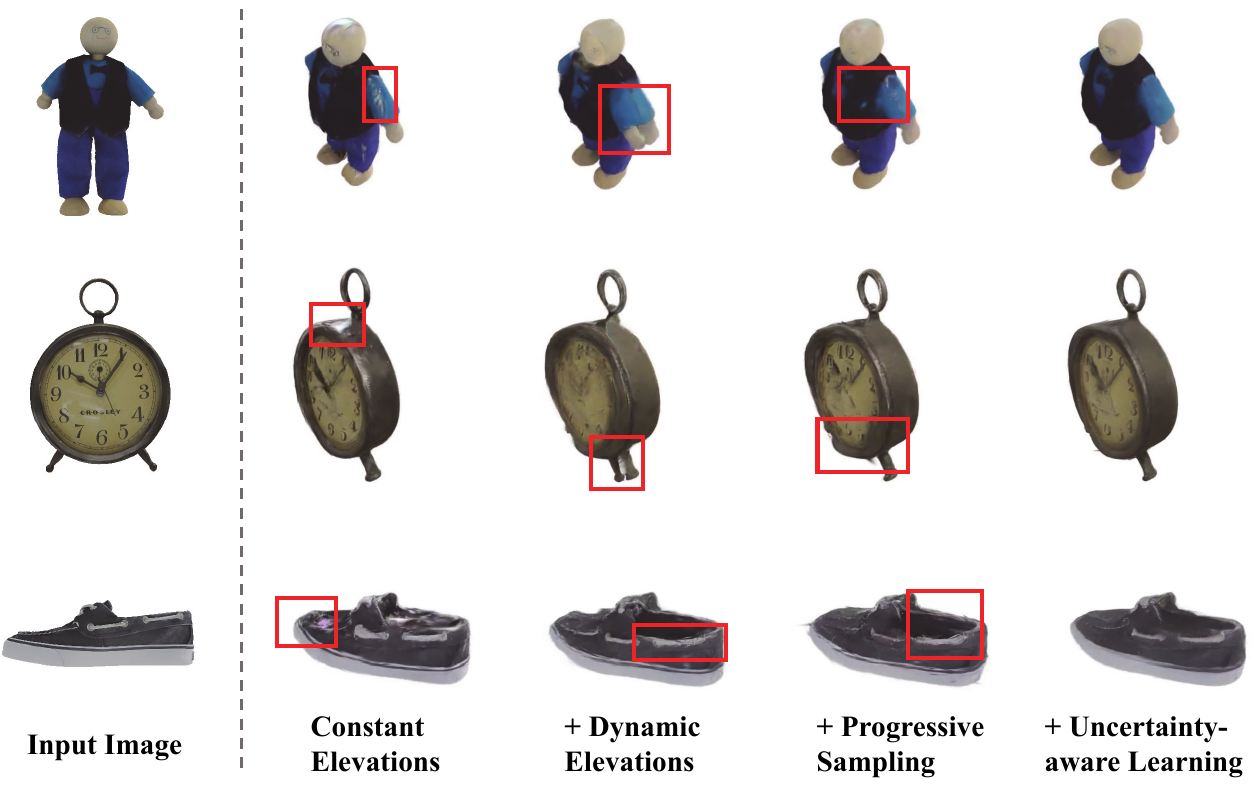}}
\caption{\textbf{Ablation analysis on the impact of progressive sampling and uncertainty-aware learning.} 
These techniques effectively mitigate artifacts, floats, and geometric deformations in inconsistent regions, resulting in visually enhanced 3D assets.
}
\label{fig:ablation_1}
\end{figure}

\begin{figure}[!t]
\centerline{\includegraphics[width=0.48\textwidth]{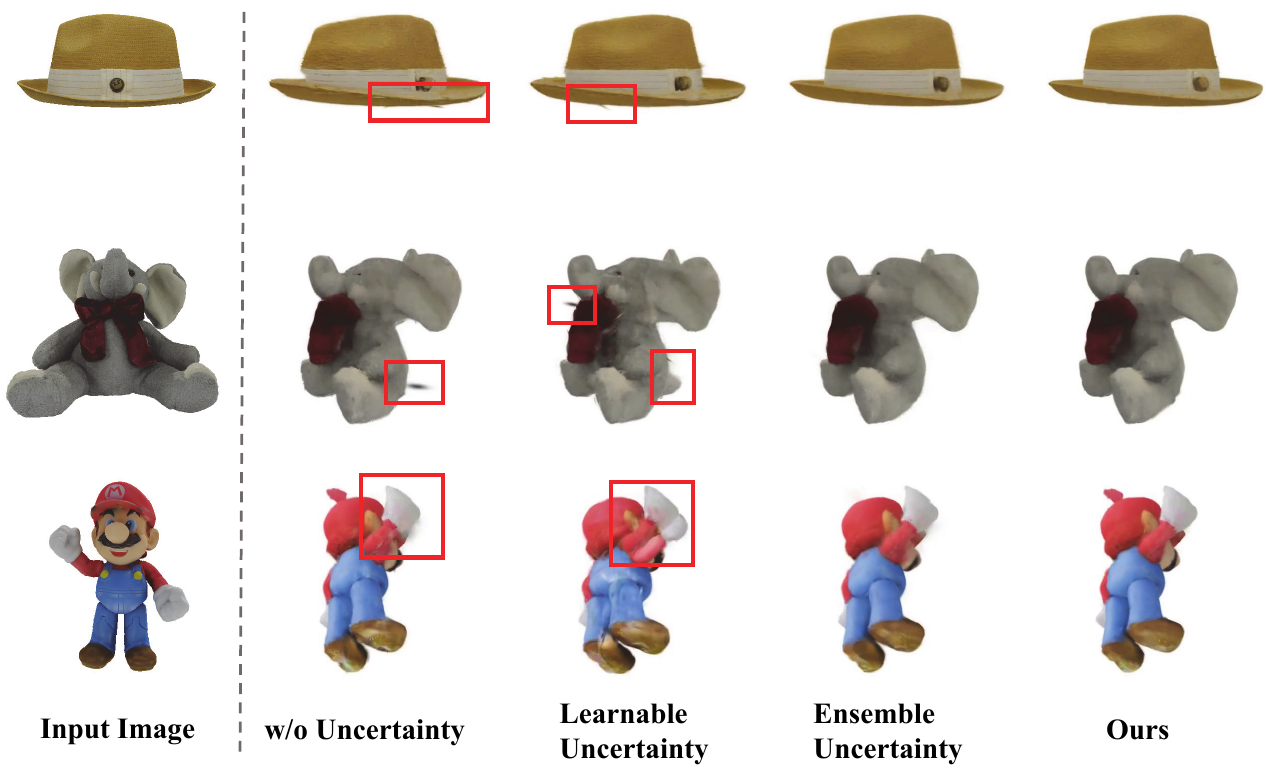}}
\caption{\textbf{Ablation analysis of the  uncertainty estimation design.} {Our method models uncertainty through the absolute difference between two concurrently optimized Gaussian models, providing stability and efficiency.}
}
\label{fig:ablation_2}
\end{figure}

\subsection{Ablation Studies and Further Discussion}

\noindent\textbf{Impact of Progressive Sampling and Uncertainty-aware Learning. }
In typical two-stage methods, video frames with fixed elevations are  generated and used as pseudo-labels to supervise the optimization of 3D assets. 
However, limited viewpoints often leave the top and bottom areas under-reconstructed, as shown in Figure \ref{fig:ablation_1}.
To address this limitation, we sample multiple videos from a wide range of viewpoints, improving coverage for top- and bottom-view perspectives.
Through progressive sampling, we start with a small number of frames to achieve an accurate initialization, and later refine texture details with the full set of frames, effectively preventing geometric deformations caused by initialization errors.
Furthermore, uncertainty-aware learning dynamically identifies inconsistent regions in the pseudo-labels and adjust the supervision strength.
Our approach effectively mitigates common issues such as artifacts and floats that often arise in prior works. 
By addressing these inconsistencies, our method ensures smoother and more accurate results, leading to visually improved 3D assets with better geometry and texture coherence.

\begin{figure}[!t]
\centerline{\includegraphics[width=0.48\textwidth]{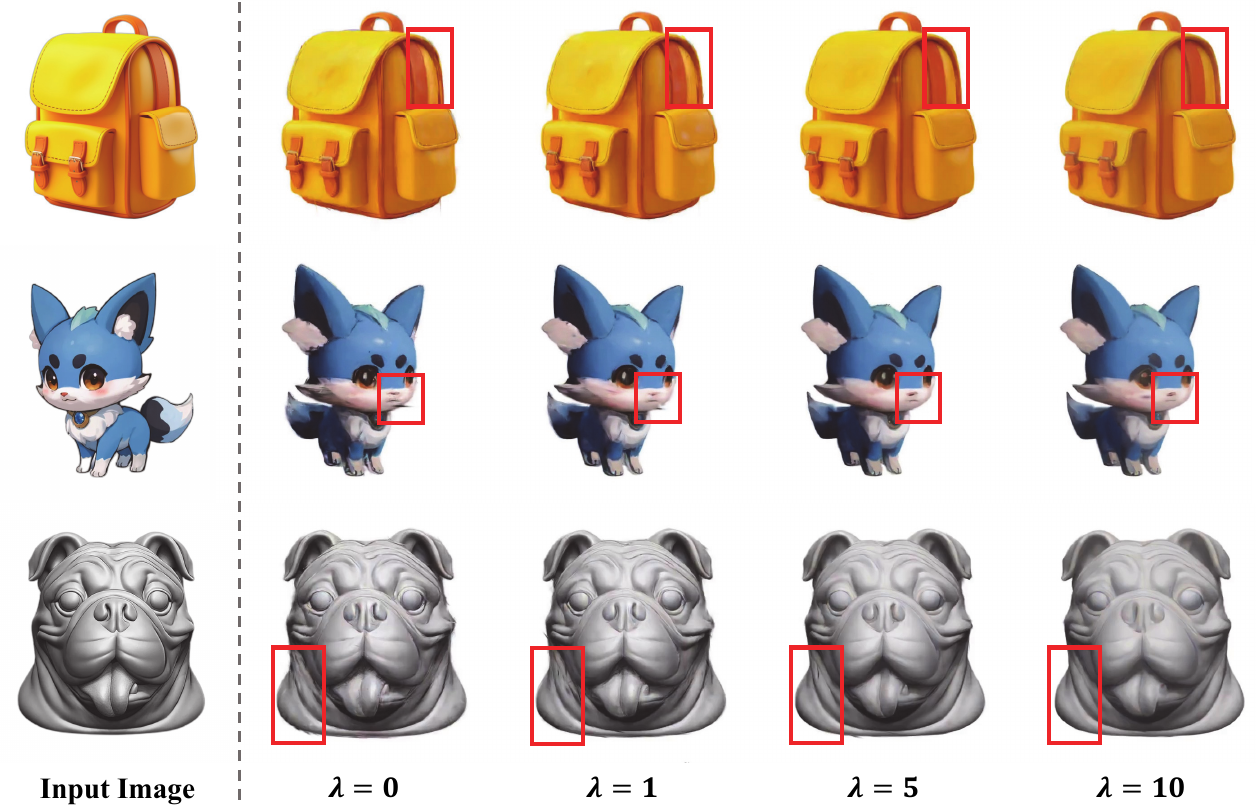}}
\caption{{\textbf{Ablation analysis of the uncertainty weight $\lambda$.} Increasing $\lambda$ reduces artifacts, but may make the generated results may become smoother and blurrier. In our experiments, $\lambda = 5$ achieves the optimal balance.}
}
\label{fig:ablation_3}
\end{figure}

\begin{figure}[!t]
\centerline{\includegraphics[width=0.48\textwidth]{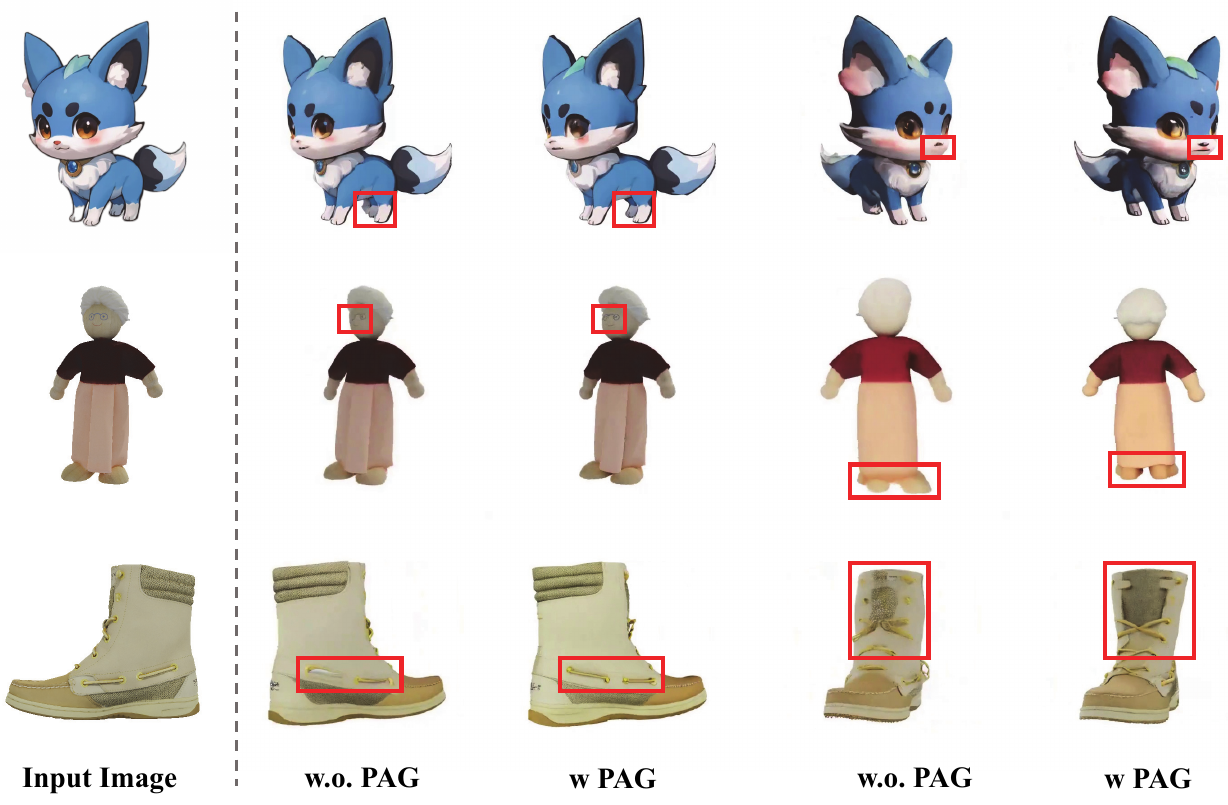}}
\caption{\textbf{Impact of Perturbed-Attention Guidance (PAG)}~\cite{pag}. PAG enhances structural coherence and texture details, improving the visual quality across the generated multi-view frames.  {In this example, red bounding boxes highlight the quality improvements introduced by PAG in specific image regions.}
}
\label{fig:suppl_ablation_pag}
\end{figure}

\noindent\textbf{Design of Uncertainty Estimation. }
We explore different approaches for uncertainty estimation, as shown in Figure \ref{fig:ablation_2}. 
Assigning a learnable variance property to 3DGS~\cite{3dgs} results in poorer performance in some cases, likely due to training instability caused by directly regressing uncertainty.
In contrast, the ensemble approach simultaneously optimizes multiple Gaussian models, averaging their predictions to generate the rendered image, with variance serving as a measure of uncertainty.
This approach effectively alleviated noise, artifacts, and floats at the edges of generated 3D assets, resulting in a smoother output. 
Our method can be seen as a simplified ensemble approach.
Experimental results show that two Gaussian models are sufficient, and their absolute difference accurately modeling uncertainty, producing similar benefits with greater efficiency.

\noindent\textbf{Impact of the Uncertainty Weight $\lambda$. }
We conduct an ablation study to assess the impact of the uncertainty weight $\lambda$, as shown in Figure \ref{fig:ablation_3}. 
When $\lambda$ is set to a low value, the pixel-wise weights in uncertainty regularization become uniform, failing to differentiate between inconsistent and consistent regions. 
This results in insufficient mitigation of artifacts and floats arising from over-reconstruction in inconsistent regions.
In contrast,  a high value for $\lambda$ substantially reduces supervision in high-uncertainty regions, effectively alleviating artifacts and floats but may lead to under-reconstruction, where inconsistent areas become overly smooth and blurry.
Our experiments show that $\lambda = 5$ strikes an optimal balance, significantly reducing artifacts while preserving texture details and preventing excessive smoothing.

\noindent\textbf{Impact of Perturbed-Attention Guidance (PAG). } 
In our approach, we employ SV3D~\cite{sv3d} to generate frames across a wide range of viewpoints. 
However, some generated samples exhibit distorted geometry or blurred textures.
To address these issues, we integrate PAG~\cite{pag} into SV3D, which enhances generation quality by guiding the denoising process away from the artificially degraded samples. 
As shown in Figure \ref{fig:suppl_ablation_pag}, the pseudo labels generated with PAG exhibit improved structure integrity and clearer, sharper textures.
\section{Conclusion}
In this work, we present a novel approach for Image-to-3D generation by incorporating uncertainty-aware learning. 
Our contributions are twofold: 
(1) we model pseudo-labels uncertainty by capturing the stochastic differences between two concurrently optimized Gaussian models; and (2) we apply uncertainty regularization, dynamically adjusting pixel-wise loss weights based on the estimated uncertainty map.
This approach effectively mitigates conflicts within the generated pseudo-frames.
By dynamically detecting inconsistencies among pseudo-labels during optimization process, our method significantly reduces artifacts and floats along the edges of the 3D assets, resulting in smoother and more accurate outputs.
Extensive experiments demonstrate the effectiveness of our approach in enhancing the visual quality of 3D generation.
{
    \small
    \bibliographystyle{ieeenat_fullname}
    \bibliography{main}
}


\end{document}